\documentclass[letterpaper]{article}
\usepackage[preprint]{aaai2027}
\usepackage{times}
\usepackage{helvet}
\usepackage{courier}
\usepackage[hyphens]{url}
\usepackage{graphicx}
\usepackage{natbib}
\usepackage{caption}
\usepackage{booktabs}
\title{Reading, Writing and Repairing\\Game-Theoretic Policies in Weight Space}
\author{
    Ancil Crayton
}
\affiliations{
   Algoverse AI Research
}

\begin{document}
\maketitle

\begin{abstract}
Can a language model work in the weight space of a policy network for a
repeated game? We put four questions to a mid-size open model, on recurrent
networks of 1{,}218 weights with exactly checkable answers. Given a population
of opponents described in English, write a policy that is optimal: its value
is a finite computation over the agent's information state, so a submitted
network has an exact gap from the optimum. Given a network's weights, report
its action after each of fifteen opponent histories; recover the state machine
it implements, judged on every opponent sequence to depth ten; or repair a
network whose output was altered in one place and say which round first goes
wrong. The model recovers the program in 62 of 66 replies, reports all fifteen
behavioral entries correctly in 34 of 36, and attains the optimum in 20 of 20
when it states a policy the evaluator compiles. Repair is the exception: 4 of
18 repaired, 5 of 18 diagnosed, with a median patch covering one of the two
entries needed.
\end{abstract}

\section{Introduction}

Recent work shows that a language model can read and write the numerical
contents of neural networks: naming the game a trained policy plays, describing
what a generative model produces, repairing a policy that fails, and writing
small networks from a game's source code \citep{wsp2026}. These demonstrations
raise a question that their evaluation cannot settle. With networks of twenty
thousand to one and a half million weights, and environments scored by returns
over sampled episodes, a policy that is right and a policy that is lucky are
hard to tell apart.

This paper asks the same questions where the answer is checkable. Our subject is
a repeated game: a $2\times2$ stage game, a population of opponents described in
English with their probabilities, and a continuation rule. The value of any
policy against that population is a finite computation and the optimum is found
by value iteration, so a submitted network has an exact value and an exact gap
from the optimum. The networks are small enough that a submitted policy can also
be compared with the network it was asked about, action by action, on every
opponent sequence to a stated depth.

Four tasks follow from that setting, differing in what the model must produce and
in what the answer is checked against. \emph{Write}: given the game, the
population and the scoring rule, and never a strategy, produce a policy.
\emph{Identify}: given a network's weights, report what it does on fifteen
opponent histories. \emph{Read}: given a network's weights, recover the state
machine it implements. \emph{Repair}: given a network that plays badly, say where
it goes wrong and patch it.

Exact scoring changes what can be claimed. A policy that is nearly optimal is
reported as nearly optimal rather than as a failure, so the distribution of
shortfalls is visible instead of collapsed into a success rate. The target is a
set, since many policies are optimal, so the task does not reward reproducing
one particular table. And instances can be constructed to order: we can guarantee
that no familiar named strategy is optimal, which is the difference between a
task that measures reasoning and a task that measures recall of a textbook.

What the four tasks show is a sharp division. Reading a network is nearly solved
at this scale: the model recovers a program in 62 of 66 replies, and 61 of those
62 are the same program under any state numbering rather than merely agreeing out
to the scoring horizon. Writing is solved in the format that removes weight
emission and costs a third of the answers in the format that requires it, so the
residual difficulty is in the emission. Repair is not solved. A two-entry fault
is located in five replies of eighteen and corrected in four, and the median
patch changes one of the two entries that need changing.

\section{Related work}

The work we replicate in a setting with exact answers is the demonstration that
language models can read and write trained weights \citep{wsp2026}. Our
contribution is not a new capability but a place to measure one, and the
measurement separates capabilities their instruments report jointly: their
experiment on naming a game is scored by keyword-matching a name, their
experiment on a generative model by keyword-matching a description, and their
network-writing experiments by return over sampled episodes.

Language models playing repeated games have been studied extensively, usually
scored by realized payoffs. \citet{coop2026} find that models defect
consistently in single-shot social dilemmas once loaded framing is removed.
\citet{nash2026} score open-weight models against Nash equilibria in one-shot
matrix games and find partial success that varies with game structure rather
than with model size. \citet{circuit2026} locate a mechanism behind one such
failure: a word-triggered circuit that suppresses equilibrium play in the named
Prisoner's Dilemma and disengages under neutral labels. That result is the
reason our target is stated as a computation over payoff numbers rather than as
a name, and our instances are labeled by whether a familiar strategy is optimal
so that the two can be separated.

The recurrent architecture and the compilation of finite-state machines into
recurrent networks follow \citet{elman1990} and \citet{omlin1996}.

\section{Task and evaluation}

\subsection{Networks and instances}

Every network is a single-layer ReLU Elman RNN with a hidden state of 32 units:
$1{,}218$ weights over the input, recurrent, and output tensors. It reads the
opponent's previous action as a one-hot vector together with a first-round flag,
so anything a policy must remember has to live in its recurrent state.

A source network implements a target policy, a finite machine driven by the
opponent's actions. We produce it three ways. \emph{Canonical} is the reference
compiler's construction, in which the machine's structure is visible in the
weights. \emph{Obfuscated} is the same function with the hidden units permuted,
rescaled and padded, so the construction no longer appears. \emph{Trained} is
gradient descent on trajectories from the target. Every source network is
verified against its target on every opponent sequence up to a stated depth
before it is used, and one that fails is dropped with its reason rather than
relabeled.

An instance for the write task is a stage game $A$ with both players' payoffs in
$[0,1]$, a population of opponent types with probabilities, and continuation
probability $\delta=0.9$. The agent is the row player and does not observe which
opponent it faces. An opponent type is a finite machine
$(\delta_{i},\mathrm{out}_{i},s^{i}_{0})$ driven by the agent's actions: in state
$s$ it plays $\mathrm{out}_{i}[s]$, and moves to $\delta_{i}[s][a]$ after the
agent plays $a$. We use eight familiar types --- always cooperate, always
defect, tit-for-tat, suspicious tit-for-tat, grim, Pavlov, tit-for-two-tats, and
an alternator --- together with seeded random machines. They are the strategies
of the iterated Prisoner's Dilemma literature, most of them entered in Axelrod's
tournaments \citep{axelrod1984}, with Pavlov from \citet{nowak1993}.

\subsection{Exact scoring}

For the write and repair tasks the value of a policy $\pi$ is
$V(\pi)=\sum_{i}p_{i}(1-\delta)\sum_{t\ge0}\delta^{t}A[a_{t},o_{t}]$ in
stage-payoff units. We report the gap $V^{*}-V(\pi)$ and its normalization by
the better constant policy,
$\rho=(V^{*}-V(\pi))/(V^{*}-V_{\mathrm{const}})$, so $\rho=0$ is optimal and
$\rho=1$ is what a policy that never adapts achieves. Success means a gap of at
most $10^{-9}$, which separates a policy that is optimal from one that differs
on a path reached with small probability.

The agent observes the opponent's actions and not its identity, so its
information state is the tuple of each type's machine state together with the
set of types still consistent with what has been observed. That space is finite
and value iteration over it gives $V^{*}$; this is the belief-state formulation
of a partially observable stochastic domain, and extracting a finite-memory
controller from the solution is the construction \citet{kaelbling1998} give for
it. The optimal policy is a machine driven by the opponent's actions, which
compiles into the recurrent architecture. An instance is admitted only if $V^{*}-V_{\mathrm{const}}$ is at
least $0.03$, so the optimum is worth reaching, and the optimal controller has
at most fifteen states. Over the twenty write instances, no familiar named
strategy is optimal on seventeen.

Identify and read are scored against the network rather than against a value.
The identify answer is the network's action after each of fifteen opponent
histories --- the empty history and every opponent sequence to length three ---
and success is exact agreement on all fifteen. Read is scored by behavioral
equivalence: the submitted machine is run against the network on every opponent
sequence to length ten, and a reply succeeds when it agrees everywhere.
Agreement out to a finite depth can in principle be a coincidence of the
horizon, so we also report whether the submitted machine is \emph{isomorphic} to
the target's minimal form, which is the right test for a program that may number
its states however it likes.

\subsection{The four families}

Table~\ref{tab:families} states what each family asks for and what it is checked
against. The answers differ in kind: a network, a table of fifteen actions, a
machine, and a patch together with the round at which the network first goes
wrong.

\begin{table}[t]
\centering
\small
\begin{tabular}{@{}lll@{}}
\toprule
Task & The model returns & Checked against \\
\midrule
write & a network or machine & exact value against $V^{*}$ \\
identify & fifteen actions & the network's behavior \\
read & a state machine & behavior to depth 10 \\
repair & a patch and a round & value, and the round \\
\bottomrule
\end{tabular}
\caption{The four tasks. Identify and read are given a network; write and
repair produce one.}
\label{tab:families}
\end{table}

\subsection{What the model is given}

For the write task the prompt states the stage game's payoffs, each opponent's
behavior in English with its probability, the continuation rule, the scoring
rule, and the network's architecture with its tensor shapes. It states that a
constant policy loses badly on every instance used, and it states the network's
input, so that anything the policy must remember has to live in the recurrent
state. It does not state what to play.

We use three reply formats for writing, chosen so that finding the policy and
expressing it can be told apart:

\begin{itemize}
\item \textbf{direct}: a JSON object holding every tensor in full.
\item \textbf{edits}: a list of numpy-style assignments applied to an all-zero
network, so only the non-zero weights appear \citep{wsp2026}.
\item \textbf{machine}: the policy as a state machine, which the evaluator
compiles into weights, removing weight emission from the task entirely.
\end{itemize}

All three are scored identically once weights exist. A reply that cannot be read
is a failure, and its status is recorded. Identify, read and repair each have one
reply format, a JSON object, and the contract defines every key in it.

\section{Baselines}

Each family is measured against the strongest thing that can be done without the
capability it is meant to measure. Every baseline below is exact, from the same
evaluator, and none calls a language model.

\textbf{Write.} The better constant policy has $\rho=1$ by construction; the
best named strategy has median $\rho=0.615$ and is exactly optimal on three
instances; evolutionary search over the weights, given 100, 1{,}000 and 10{,}000
exact evaluator queries from a zero start, has median $\rho$ of 0.97, 0.82 and
0.83; and the reference compiler is optimal. The search result gives the task
its point: blind optimization of the same weights, with ten thousand exact
evaluations of them, barely improves on not adapting.

\textbf{Identify.} Simple statistics of the weights recover the state count and
nothing further. Counting the hidden units that both read a signal and send one
gives $2n+1$ for an $n$-state machine on canonical weights, and the obfuscator
moves those counts without erasing the ordering. But one fingerprint of
twenty-two summary statistics --- the active-unit count, the width, and four
per-tensor statistics for each of five tensors --- covers grim, tit-for-tat and
a shuffled control together, so the statistic cannot separate two targets with
the same state count, or either from a network whose weights mean nothing.

\textbf{Read.} The strongest procedure without weight access is to ask the
network what it does. A hill-climb over machines, given $N$ probes of the
network's behavior and told the state count, recovers the program from 0 of 8
machines at 0 probes, 1 of 8 at 4, 5 of 8 at 16, 6 of 8 at 64 and 8 of 8 at 256.

That hill-climb is weaker than the established method and we report it as a
floor. Extracting a finite automaton from a recurrent network is a solved
problem: \citet{weiss2018} treat the network as a membership oracle and run
Angluin's learner \citep{angluin1987}, which needs polynomially many queries
given an equivalence oracle. Our search takes no equivalence oracle, because a
deployment has none, so 256 probes is what behavior buys under that constraint
and not the best that can be done. The comparison the family turns on is
therefore against a strong query-based learner, not a weak one.

\textbf{Repair.} The write family's search, started from the flawed network
rather than from nothing, repairs 0 of 6 at $10^2$ and $10^3$ evaluations and 1
of 6 at $10^4$, while rewriting every weight. The reference patch changes two
entries out of $1{,}218$.

\section{Results}

Table~\ref{tab:results} collects the four families. Every reply is from
\texttt{deepseek-v4.1-flash-1017}, a mid-size open model, called through
OpenRouter at reasoning effort \texttt{high}, blind: one reply per call, no
tools, no feedback, three replicates per cell. Success is per reply and the
intervals are Wilson 95\%. A reply that cannot be read is a failure.

\begin{table}[t]
\centering
\footnotesize
\begin{tabular}{@{}llrc@{}}
\toprule
Task & Measure & Result & 95\% CI \\
\midrule
write & machine format & 20/20 \ 1.000 & [0.839, 1.000] \\
 & sparse weight list & 18/20 \ 0.900 & [0.699, 0.972] \\
 & every weight in full & 13/20 \ 0.650 & [0.433, 0.819] \\
\midrule
identify & table exact & 34/36 \ 0.944 & [0.819, 0.985] \\
 & individual actions & 528/540 & --- \\
 & name (secondary) & 30/36 \ 0.833 & [0.681, 0.921] \\
\midrule
read & program recovered & 62/66 \ 0.939 & [0.854, 0.976] \\
 & isomorphic & 61/66 \ 0.924 & [0.835, 0.967] \\
 & state count right & 55/66 \ 0.833 & [0.726, 0.904] \\
 & window only & 0/66 \ 0.000 & [0.000, 0.055] \\
\midrule
repair & repaired & 4/18 \ 0.222 & [0.090, 0.452] \\
 & diagnosed & 5/18 \ 0.278 & [0.125, 0.509] \\
 & both & 2/18 \ 0.111 & [0.031, 0.328] \\
 & no harm & 15/18 \ 0.833 & [0.608, 0.942] \\
\bottomrule
\end{tabular}
\caption{The four tasks, one model, exact scoring throughout. Intervals are
Wilson 95\% on the reply count; the action total is an integer and has none.}
\label{tab:results}
\end{table}

\subsection{Writing: the difficulty is in the emission}

How much the model achieves depends on how much it must write. Stating the
policy as a state machine, which the evaluator compiles, attains the optimum on
every instance. Writing the weights as a sparse list of assignments attains it
on eighteen of twenty, and writing every weight in full on thirteen. Median
regret is zero in all three, so the losses are not a general drift away from the
optimum but whole replies that land far from it.

A result of this kind is only interesting if it is not recognition. Seventeen of
the twenty instances are ones on which no familiar strategy is optimal, so on
those no memorized policy can be the answer.
Table~\ref{tab:write} gives the split. All three formats are exactly optimal on
all three instances where a familiar strategy is optimal, which is the easy
half; on the seventeen where none is, the formats separate: ten, fifteen and
seventeen. The machine format is exactly optimal on every instance in the
stratum that recall cannot explain.

That ordering is the write result. If the model were recognizing strategies, its
success would fall away where the strategies do not apply; instead it is
complete there in the format that carries no weight emission, and the shortfall
that remains in the other two is a shortfall in writing the answer rather than
in finding it.

\begin{table}[t]
\centering
\small
\begin{tabular}{@{}lccrr@{}}
\toprule
Format & Solved & Rate & named opt. & none opt. \\
\midrule
direct  & 13/20 & 0.650 & 3/3 & 10/17 \\
edits   & 18/20 & 0.900 & 3/3 & 15/17 \\
machine & 20/20 & 1.000 & 3/3 & 17/17 \\
\midrule
best named strategy & 3/20 & 0.150 & \multicolumn{2}{c}{median $\rho=0.615$} \\
search, $10^4$ queries & 0/20 & 0.000 & \multicolumn{2}{c}{median $\rho=0.826$} \\
\bottomrule
\end{tabular}
\caption{Writing, by format and by whether a familiar named strategy happens to
be optimal. The last two columns are the stratum recall cannot explain.}
\label{tab:write}
\end{table}

\subsection{Identify: reading a network's behavior}

The model reports the network's action after each of fifteen opponent histories.
It is exact on thirty-four of thirty-six replies, which is five hundred and
twenty-eight individual actions out of five hundred and forty; the chance rate
for one reply is $2^{-15}$. The successes include the cases where recall cannot
be the explanation: eighteen of eighteen on named targets, ten of twelve on
unnamed random machines and six of six on networks whose weights were shuffled,
both of the latter having the refusal as their correct answer.

Table~\ref{tab:identify} splits the result by how the weights came to implement
the target. The same machines are used in both arms, so the two rates are
comparable. Canonical weights are read perfectly; the obfuscated arm loses two
replies, and both are the \emph{same machine} answering with the \emph{same}
fifteen actions \emph{shifted one position}: the model read the weights and
misplaced the answer, so the two misses are one ordering slip rather than two
failures of reading. On every other obfuscated instance the answer is exact.

The optional name is a different matter and is reported beside the table rather
than folded into it. It is correct on thirty of thirty-six, and the misses are
cases where the model described what it saw instead of naming a standard
strategy. That the table is right where the name is wrong is the expected
direction: producing fifteen actions is simulation, and naming a strategy is
recognition from a fixed vocabulary.

\begin{table}[t]
\centering
\small
\begin{tabular}{@{}lccc@{}}
\toprule
Weights & Table exact & Rate & Name \\
\midrule
canonical  & 21/21 & 1.000 & 18/21 \\
obfuscated & 13/15 & 0.867 & 12/15 \\
\bottomrule
\end{tabular}
\caption{Identify, by whether the reference construction is visible in the
weights. The same machines appear in both arms.}
\label{tab:identify}
\end{table}

\subsection{Read: recovering the program}

Scored by behavioral equivalence to depth ten, seven rounds past the window the
identify answer covers, the model recovers the program on sixty-two of sixty-six
replies. Sixty-one of those are isomorphic to the target, so the score is not
being carried by coincidence at a short horizon: asked about tit-for-tat the
model returns the target machine entry for entry, with no state relabeling.

The failure mode the family was designed around --- a machine that matches over
the window and diverges after it --- did not occur on any reply. The one
substantive failure is the opposite: a machine of thirty-six states for a
two-state target, right for two rounds and wrong afterwards, which is why the
right state count is reported separately from isomorphism.

Table~\ref{tab:read} splits the result by provenance. The arms do not hold the
same machines, because training did not reach exactness for the two largest,
so the comparison is computed on the six machines present in every arm.
Obfuscation costs nothing measurable, and canonical weights are the worst of the
three arms by two replies out of eighteen; with that many failures the result
supports no ordering between the arms.

\begin{table}[t]
\centering
\small
\begin{tabular}{@{}lccc@{}}
\toprule
Provenance & Recovered & Rate & all machines \\
\midrule
canonical  & 16/18 & 0.889 & 22/24 \\
obfuscated & 18/18 & 1.000 & 22/24 \\
trained    & 18/18 & 1.000 & 18/18 \\
\bottomrule
\end{tabular}
\caption{Read, by how the weights came to implement the target. The first two
columns are the six machines present in every arm; the last includes machines
only the compiled arms have.}
\label{tab:read}
\end{table}

\subsection{Repair: locating a fault and correcting it}

The network handed over is a controller with one output changed, and the change
is required to cost at least a tenth of the gap between the optimum and the best
constant policy. Two things are scored and neither is folded into the other: the
patch, by the value of the result, and the diagnosis, the shortest opponent
sequence on which the network plays wrongly.

The model repairs four of eighteen replies and diagnoses five, and the two sets
are not nested: three replies located the fault and failed to fix it, two did
both, and no reply repaired a network it had misdiagnosed. So the repair rate is
not a proxy for the diagnosis rate, and the difference between them is the
measurement the family exists to make.

The patch sizes explain most of the failures. The planted change moves one
hidden unit's contribution from one action to another, which is two entries of
the output weights: one set to zero and one to a large positive value. The
model's median patch is \emph{one} entry. A model that believes the fault is a
single weight corrects one of the two and leaves the network short of the
optimum.

The wording is part of the explanation and we tested it. Rewriting "changed in
one output" as "changed in one place: the action the controller plays in one
state is not the action a correct controller would play there" gives the same
information without the numeral. It moves the median patch from one entry to two
and the number of replies that change fewer entries than the fault needs from
fourteen to nine, so the numeral was suppressing the edit. It does not move the
repair rate: four of eighteen against two, on eighteen replies each, a difference
of two replies. What it does change is the kind of repair. Under the original
wording all four successes were single edits that reached a different optimal
policy; under the second, one of the two was the two-entry correction the fault
actually requires. Removing a misleading numeral makes the faithful edit possible
and does not make it likely, because choosing the two values is the hard part.

Four of the reported repairs used a patch of one entry where the reference patch
needs two. They succeeded, because the optimal policy is a set and a different
single edit can reach it. This is why patches are never compared against the
reference patch, and it is not a hypothetical: a rule that matched entries would
have failed four correct repairs. Two further replies changed both entries to
values that leave the network worse than it started, which is why no-harm is
reported beside the repair rate.

\subsection{Where the failures fall}

Three of the 120 replies across the three new families were lost to truncation,
where the model reasons at length and the output budget is exhausted with no
answer emitted. This is a property of the generation budget rather than of the
task, and it is measured rather than assumed: the truncated replies reasoned for
their entire budget, while the completed ones emitted answers of twenty to forty
tokens after several thousand tokens of reasoning, so the answer was never the
expensive part.

The remaining failures are readable and wrong, and they are the failures the
tasks are designed to measure. Three are worth naming. Two identify replies on
one obfuscated instance return the target's fifteen actions shifted one
position, which is a misplaced answer rather than a misread one. One read reply
is a machine of thirty-six states where two suffice. Two repair replies make the
network worse than the one they were given, which is why no-harm is reported.

\subsection{Relation to the published results}

Table~\ref{tab:published} puts this paper's four measures beside those published
by \citet{wsp2026} for the experiments they replicate, and Table~\ref{tab:setup}
states what each study put in front of the model and what the answer was graded
against.

\begin{table*}[t]
\centering
\footnotesize
\begin{tabular}{@{}llp{0.34\textwidth}p{0.34\textwidth}@{}}
\toprule
Experiment & & \citet{wsp2026} & This paper \\
\midrule
identify & subject & four PPO networks, one per named game,
17{,}253 weights & twelve networks in two arms, $1{,}218$ weights each \\
 & answer & five free-text fields & fifteen actions and an optional name \\
 & graded by & keyword match on the game name in one field & exact agreement
on all fifteen, then the optional name \\
\midrule
describe & subject & three diffusion models, 23{,}824 weights & eight
machines in three provenances, $1{,}218$ weights \\
 & answer & a description and a $12\times12$ rendering & a state machine \\
 & graded by & keyword match on the described shape & behavioral equivalence
to depth ten, and isomorphism \\
\midrule
repair & subject & one actor, 171{,}664 weights, failing every episode & six
instances, $1{,}218$ weights, one output changed \\
 & evidence & its score and a recorded episode & the weights and the population
\\
 & graded by & success over 1{,}024 episodes & exact value against $V^{*}$, and
the round of the first wrong action \\
\midrule
write & subject & three MinAtar games, from the game's source code, 20{,}486
weights & twenty instances, from the payoffs, $1{,}218$ weights \\
 & answer & weights, as edit statements & weights, in three reply formats \\
 & graded by & mean return over 1{,}000 episodes & exact value against $V^{*}$
\\
\bottomrule
\end{tabular}
\caption{What \citet{wsp2026} put in front of the model and what this paper puts
in front of it, and what each answer is graded against. The subjects differ by an
order of magnitude in size and the grading differs in kind in every row.}
\label{tab:setup}
\end{table*}

\begin{table}[t]
\centering
\small
\begin{tabular}{@{}lll@{}}
\toprule
Experiment & \citet{wsp2026} & This paper \\
\midrule
identify a game & 4/4 & 34/36 \ 0.944 \\
describe a generator & 3/3 & 62/66 \ 0.939 \\
repair a failing policy & 1{,}024/1{,}024 & 4/18 \ 0.222 \\
write from scratch & 517.4 vs 25.3 & 20/20 \\
\bottomrule
\end{tabular}
\caption{This paper's measures against those published by \citet{wsp2026}. The
scoring differs in every row, and the last row compares a sampled return with an
exact success rate.}
\label{tab:published}
\end{table}

Their measures have no ceiling and ours do: a sampled
return is unbounded, so a hand-written network can exceed a network trained for
five hundred million steps, as the published Breakout network does, and a return
can never establish that a policy is optimal. The measures in this paper are
scored against a computed optimum, where being optimal is the most that can be
said. Their first two experiments
are graded by keyword-matching a free-text field, so four of four and three of
three are counts of cases rather than rates, and neither can separate a model
that reads a network from one that recognizes a name; that separation is what
our controls and our exact answer formats exist to make.

The repair row is the largest divergence and it is not a like-for-like
comparison. Their prompt supplies the failing policy's score and a recorded
episode; ours supplies neither, because our family's claim is that the fault is
found by reading the weights. Their failure is total and ours costs a tenth of
a span. Their repair adds a sub-path through a few units, 3.5\% of the weights,
while ours substitutes two entries whose values the model cannot know.

The write row differs in subject rather than in evidence. They write
20{,}486-weight networks for MinAtar games from the game's source code and are
scored by mean return over a thousand episodes; we write $1{,}218$-weight
networks for a $2\times2$ repeated game from its payoffs and are scored against a
computed optimum. Their number is a return that a trained network can be beaten
on and ours is a success rate whose ceiling is one, so the two cannot be ranked
against each other. What our row adds is a decomposition their instrument cannot
make, because every one of their runs writes weights: the same policy stated as a
machine the evaluator compiles is optimal on every instance, while writing every
weight in full is optimal on thirteen of twenty. On the same instances and the
same policy, what separates the two is the length and structure of the text the
model must produce.

One published result is one this paper's instrument could not have produced:
their model named all three generated shapes correctly and drew all three
wrongly, an outline where the true samples are filled. Two measurements made
that visible. Our exact scoring would have recorded three of three and seen
nothing, which is the argument for reporting identify's optional name beside its
table rather than folding the two together.

\section{Discussion}

The four tasks give an ordering that is not the one a reader might expect. At
this scale, recovering a program from weights is nearly solved, and recovering a
network's behavior is solved: sixty-one of sixty-six read replies are the same
program under any state numbering, and five hundred and twenty-eight of five
hundred and forty identify actions are right. That is not recognition of a name. It is simulation of a
network from its parameters, repeated.

Writing is solved in the format that removes weight emission and costs between a
quarter and a third of the answers in the formats that require it. The three
formats pose the same question and differ only in how the answer must be given,
and they score 0.65, 0.90 and 1.00. When the answer is a state machine that the
evaluator compiles, the model is exactly optimal on every instance, including all
seventeen where no familiar strategy is optimal. Since the instances are the same
and the policy is the same, what separates the formats is the length and the
structure of the text the model must produce.

Repair is different in kind. The fault is two entries out of $1{,}218$, the
optimum is computable from the description the model already has, and ten
thousand exact evaluations of the same network repair one instance in six. A
model that located the fault would need to change two numbers. It locates the
fault in five replies of eighteen and changes one entry in the median reply, and
the gap between those two facts is where a repair fails.

Work of this kind reports a single accuracy figure for weight writing, and that
figure conflates two things an experiment can separate: whether the model knows
what policy to implement, and whether it can write that policy down. Our machine
format removes the second question and the first becomes visible. The same
separation applied to repair splits the rate into a diagnosis and a fix, and the
two come apart in the data rather than only in the analysis.

\section{Limitations}

The stage games are $2\times2$ and the networks hold $1{,}218$ weights, some four
orders of magnitude below the networks in the work we compare with. Scale is a
real question: the reply formats accommodate larger networks, and a
thirty-six-state machine was produced for a two-state target, but nothing here
shows that the results survive at a million weights.

The identify family runs on canonical and obfuscated weights but not on trained
ones, because the machines it uses are drawn per instance rather than reused from
the read family. Canonical weights make the reference compiler's construction
visible, and a simple statistic recovers the state count from them, though not
the strategy; the obfuscated arm is the control for that, and it loses two
replies of fifteen.

The repair measurement is eighteen replies over six instances, and its interval
is wide. The per-instance pattern is more informative than the rate: three
instances fail with the same one-entry patch in all three replicates, which is a
statement the interval does not capture.

The read arms do not hold the same machines, because training did not reach
exactness for the two largest. The provenance comparison is therefore computed
on the six machines every arm contains, and it rests on eighteen replies per arm.

The measurement is one model, one seed stream, three replicates. The intervals
are wide and the differences between the write formats are not distinguishable
at this size. The claim the data support is that the model achieves exact
success on a substantial fraction of instances, and that its failures are
concentrated in emission and in patching rather than in comprehension.

Finally, our success criterion is exactness. Some replies fail it while trailing
the optimum by a tenth of a percent, and we report those as near-optimal rather
than as failures; the exact figure is the stricter of the two and should be read
as such.

\section*{Code and data availability}

The analysis scripts, the stored model responses, and the scripts that
regenerate every number in this paper from them are available from the author on
request. Every score reported here is re-derived from a stored response rather
than read from a previously computed summary, so the tables can be rebuilt
without any further model calls.

\bibliography{refs}

\end{document}